\pdfoutput=1

\documentclass[11pt]{article}

\usepackage[final]{acl}

\usepackage{times}
\usepackage{latexsym}

\usepackage[T1]{fontenc}

\usepackage[utf8]{inputenc}

\usepackage{microtype}

\usepackage[utf8]{inputenc} 
\usepackage[T1]{fontenc}    
\usepackage{hyperref}       
\usepackage{url}            
\usepackage{booktabs}       
\usepackage{amsfonts}       
\usepackage{nicefrac}       
\usepackage{microtype}      
\usepackage{xcolor}         
\usepackage{enumitem}
\usepackage{graphicx}
\usepackage{amsmath}
\usepackage{multirow} 
\usepackage{algorithm}
\usepackage{algorithmic}

\usepackage{inconsolata}

\usepackage{graphicx}

\title{MemeArena: Automating Context-Aware Unbiased Evaluation of Harmfulness Understanding for Multimodal Large Language Models}

\author{Zixin Chen$^{\heartsuit}$\thanks{\; Equal contribution.}, Hongzhan Lin$^{\spadesuit *}$\thanks{\; Corresponding authors.}, Kaixin Li$^{\diamondsuit}$, Ziyang Luo$^{\spadesuit}$, \textbf{Yayue Deng}$^{\heartsuit}$, \textbf{Jing Ma}$^{\spadesuit \dagger}$ \\ 
        $^{\spadesuit}$Hong Kong Baptist University, Hong Kong\\
        $^{\heartsuit}$Beijing University of Posts and Telecommunications, China\\
        $^{\diamondsuit}$National University of Singapore, Singapore\\
        \texttt{\{mailboxforvicky\}@bupt.edu.cn},
        \texttt{\{cshzlin,majing\}@comp.hkbu.edu.hk}}

\begin{document}
\maketitle
\begin{abstract}
The proliferation of memes on social media necessitates the capabilities of multimodal Large Language Models (mLLMs) to effectively understand multimodal harmfulness. 
Existing evaluation approaches predominantly focus on mLLMs' detection accuracy for binary classification tasks, which often fail to reflect the in-depth interpretive nuance of harmfulness across diverse contexts. 
In this paper, we propose MemeArena, an agent-based arena-style evaluation framework that provides a context-aware and unbiased assessment for mLLMs' understanding of multimodal harmfulness.
Specifically, MemeArena simulates diverse interpretive contexts to formulate evaluation tasks that elicit perspective-specific analyses from mLLMs. By integrating varied viewpoints and reaching consensus among evaluators, it enables fair and unbiased comparisons of mLLMs’ abilities to interpret multimodal harmfulness.
Extensive experiments demonstrate that our framework effectively reduces the evaluation biases of judge agents, with judgment results closely aligning with human preferences, offering valuable insights into reliable and comprehensive mLLM evaluations in multimodal harmfulness understanding.
Our code and data are publicly available at \url{https://github.com/Lbotirx/MemeArena}.
\end{abstract}

\section{Introduction}
The boom of social media has led to the advent of a new prevalent form of multimodal entity: Meme. Typically comprising combinations of visual elements and short texts, memes can be easily shared and quickly spread across diverse communities on various social platforms. 
While often humorous or sarcastic~\cite{hessel2023androids, chen2024cofipara}, memes can also convey harmful or hateful multimodal messages when interpreted through different cultural or social lenses.

A widely accepted definition of harmful memes\footnote{\color{red}\textbf{Disclaimer:} This paper contains content that may be disturbing to some readers.} is “multimodal units consisting of an image and accompanying text that have the potential to cause harm to an individual, an organization, a community, or society in general”~\cite{sharma2022detecting}. Recent works~\cite{lin2023beneath,cao2023pro,kumari2024m3hop,lin2024explainable} have been increasingly integrating multimodal Large Language Models (mLLMs) into harmful meme detection, leveraging their extensive background knowledge to address issues related to meme-based social abuse~\cite{ kiela2020hateful,pramanick2021detecting, fersini2022semeval}. This trend has highlighted the importance of systematically assessing the reasoning capabilities of mLLMs in understanding meme harmfulness, with the aim of advancing their deployment for online trust and safety applications.

\begin{figure*}[t!]
    \centering    
    \includegraphics[width=\textwidth]{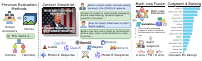} 
    \vspace{-0.9cm}
    \caption{An overview of previous evaluation paradigms and our proposed MemeArena framework. 
}
    \label{fig:overview}
    \vspace{-0.3cm}
\end{figure*}

Existing evaluation approaches~\cite{lin2025goat} typically judge the mLLM's capability of multimodal harmfulness understanding by emphasizing detection accuracy in a binary classification manner.
However, such simple task design and judgment metric fall short of providing a comprehensive mLLM evaluation, as they typically gauge harmfulness based solely on predefined labels, thereby limiting the depth and nuance of meme interpretation.
In particular, memes encountered in real-world contexts are inherently open to multi-perspective interpretations, as the public with different cultural, social, or political backgrounds may perceive the same meme in vastly different ways. 
For instance, 
as shown in Figure~\ref{fig:overview}, the meme features Donald Trump questioning the CDC’s mask recommendation with an absurd justification. Policy experts may interpret its harmfulness in terms of public health politicization and political polarization, while less politically informed viewers may see it simply as a joke that downplays the pandemic’s severity.
These interpretive differences underscore the importance of developing open-ended evaluation tasks that consider specific contexts and perspectives in understanding harmfulness, while traditional closed-ended benchmarks~\cite{kiela2020hateful, pramanick2021detecting} often struggle to accommodate the nuance in a scalable and reliable manner.
On the other hand, the judgment of mLLMs' performances also faces the challenge presented by such contextual differences, as the evaluators themselves may hold this divergence in their conceptions of harmfulness as well, leading to subjective evaluation results.
Accordingly, a fair and unbiased assessment requires the incorporation of opinions from a broad and diverse group of evaluators to account for these differences in understanding.
A potential solution is to leverage platforms like Chatbot Arena~\cite{chiang2024chatbot}, which collects crowdsourced annotations from a wide range of users and utilizes pairwise comparisons to enable nuanced and objective judgments of LLMs' capabilities. Yet, this approach requires extensive efforts for data preparation, and thus becomes extremely costly when evaluations are conducted at scale~\cite{zheng2023judging}.

To address these challenges, we propose to develop an automated arena-like framework, by incorporating a diverse range of interpretative perspectives and evaluator opinions, to facilitate comprehensive and reliable evaluations on mLLMs' understanding of multimodal harmfulness.
Specifically, in this paper, we design the evaluation framework with the following key principles: 
1) The framework should facilitate evaluation tasks from a wide spectrum of perspectives in interpretations. As harmfulness can be perceived in varied ways~\cite{huang2023chatgpt}, 
we propose to simulate a varying set of scenarios for each meme, and enable perspective-specific analysis to thoroughly evaluate mLLMs' context-aware comprehension of harmfulness.
2) The framework needs to conduct impartial assessments that integrate diverse evaluator opinions into the judgments of mLLMs' performances.
To reduce individual bias, we draw on the concept of collective intelligence~\cite{leimeister2010collective}, aligning judgments among a group of varied evaluators through establishing consensus in harmfulness understanding.
3) The framework should also automate the whole evaluation process to promote effective and scalable evaluations. Thus we propose to audit mLLMs modularly in an agentic way~\cite{park2023generative, gu2024survey}, avoiding the heavy reliance on costly manual efforts.

To this end, we introduce \textbf{MemeArena}, a novel agent-driven evaluation framework that assesses mLLMs' capabilities to understand context-aware multimodal harmfulness in an arena-style fashion. Our framework includes three stages:
1) Context Simulation \& Task Formulation: We first leverage agents to simulate interpretive contexts by generating profile demographics with diverse social backgrounds. Based on these profile-derived interpretive contexts, we formulate context-specific tasks that enable varied, perspective-specific harmfulness analyses for memes.
2) Multi-view Fusion: With the context-specific tasks, target mLLMs are instructed to generate responses to interpret each meme from multiple perspectives, which are then aggregated by a group of judge agents through a multi-view fusion process. This procedure adaptively integrates insightful model responses as well as the opinions of diverse judges to construct comprehensive guidelines, capturing aligned values among judges in harmfulness understanding.
3) Judgment \& Ranking: Utilizing the guidelines as references, the responses of target models are then compared and ranked based on their reasoning quality in an arena-like manner, resulting in fair and unbiased judgments of target mLLMs’ performances. 

Our contributions can be summarized as follows:
\begin{itemize}[leftmargin=*,nosep]
\item  To the best of our knowledge, this is the first work to automatically evaluate mLLMs' context-aware harmfulness understanding of multimodal meme content through an analytical lens, on context-centric interpretations of multimodal harmfulness across diverse contextual perspectives. 
\item We present \textbf{MemeArena}, an agent-based evaluation framework that provides unbiased assessments on mLLMs’ multimodal understanding of meme harmfulness. Our method automates the process of task design and value-aligned judgments, to facilitate fair and comprehensive evaluations that account for the complex and subjective nature of harmfulness interpretations.
\item Experimental results show that our method substantially reduces bias in evaluating mLLMs’ understanding of meme harmfulness. Judgments from diverse agents exhibit strong consistency and closely align with human preferences, offering valuable insights for reliable mLLM audits.
\end{itemize}

\section{Methodology}
\label{headings}

\subsection{Overview} 
\paragraph{Problem Statement} Multimodal harmfulness understanding in memes focuses on interpreting how and why the multimodal content of memes may contribute to harmful meanings.
In our proposed MemeArena, we aim to conduct a comprehensive, context-aware, and unbiased arena-style evaluation that automatically assesses mLLMs’ abilities to interpret multimodal harmfulness in memes from diverse socio-cultural perspectives.
Specifically, on a set $\mathcal{M}$ of harmful meme data, we evaluate a group of target mLLMs $\{\mathcal{T}_1,...,\mathcal{T}_n\}$, with an evaluator panel $\mathbb{J}$ consisting of judge agents as follows:
\begin{equation}
\mathcal{R} = \text{MemeArena}(\mathcal{T}_1,...,\mathcal{T}_n|\mathbb{J},\mathcal{M},\mathcal{C}_{\mathcal{M}}),
\end{equation}
where $\mathcal{C}_{\mathcal{M}}$ is a set of interpretive contexts designed to simulate diverse perspectives relevant to the interpretations of memes, 
the evaluator panel $\mathbb{J} = \{{J_1, \dots, J_k}\}$ consists of $k$ judge agents, and $\mathcal{R}$ denotes the context-aware evaluation outcomes in the form of a ranking that reflects the relative capabilities of the $n$ ($n>k$) target mLLMs in understanding multimodal harmfulness of memes.

Given the complex and subjective nature of multimodal harmfulness interpretations, holistic and inclusive evaluations necessitate the judge agent's thorough understanding of harmfulness to ensure fair and unbiased evaluation results.
Our core idea is to probe the target mLLM's understanding capacities of multimodal harmfulness from multiple analytical context-centric perspectives and synthesize diverse evaluator opinions to forge an unbiased consensus. This promotes a comprehensive understanding of meme content and enables fair comparison and ranking of mLLM performance in an arena-like manner. 
As illustrated in Figure~\ref{fig:overview}, our framework consists of three stages: Context Simulation \& Task Formulation (\S\ref{sec:context_sim}), Multi-view Fusion (\S\ref{sec:preference_fuse}), and Judgment \& Ranking (\S\ref{sec:judgement_rank}). 

\subsection{Context Simulation \& Task Formulation}\label{sec:context_sim}
The meme harmfulness can be perceived differently in a variety of contexts, as the multimodal humor of memes typically stems from specific sociocultural narratives that do not translate uniformly among their audiences~\cite{huang2023chatgpt}. 
To capture this diversity in harmfulness understanding, we aim to design context-specific tasks from the standpoint of individuals situated in varied background settings, reflecting how different people might interpret the same meme through context-sensitive lenses.

\begin{figure*}[t!]
    \centering    \includegraphics[width=\textwidth]{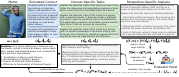} 
    \vspace{-0.8cm}
    \caption{An illustration of context simulation, task formulation and multi-view fusion. We first simulate diverse interpretive contexts, and formulate context-specific tasks to enable target mLLMs' perspective-specific analyses, which are then iteratively and adaptively refined by the evaluator panel into a multi-view fused guideline.
}
    \label{fig:method}
    \vspace{-0.3cm}
\end{figure*}

As demonstrated in Figure~\ref{fig:method}, given a multimodal meme $m = \{i,t\} \in \mathcal{M}$ consisting of a meme image $i$ embedded with a meme text $t$, we first simulate a range of interpretive contexts by identifying specific individuals or groups with different levels of relevance related to the meme content. To reflect and ensure the diversity of interpretive contexts~\cite{ge2024scaling}, we characterize concrete and differentiated demographic profiles into three distinct types:
1) someone with a background highly relevant to the meme; 2) someone with a moderately relevant background related to the meme; 3) someone with a completely unrelated background who only encounters the meme by chance.
Consequently, the simulated context is denoted as
$\{{c^1_m, c^2_m, c^3_m}\}$, with each $c_m$ serving as a unique interpretive lens grounded in the socio-cultural background of specific individuals or groups. 
Combined with the meme $m$, we then formulate a context-specific task $x_m$ for each context $c_m$, which requires the mLLMs to discern harmfulness from a context-specific perspective, designed to elicit a different nuance of the meme’s interpretations. After that, each target mLLM $\mathcal{T}$ is then instructed to generate an analysis set $y(\mathcal{T},m)$ to decipher the harmfulness separately under these simulated contexts:
\begin{equation}
    y(\mathcal{T},m) = \{\mathcal{T}(x^1_m),\mathcal{T}(x^2_m),\mathcal{T}(x^3_m)\}.
\end{equation}
Since understanding multimodal harmfulness in memes requires mLLMs not only to analyze the superficial content of the image and text, but also to perform deeper reasoning grounded in commonsense~\cite{lin2023beneath}, we decompose the analysis task into two sub-dimensions~\cite{jiang2025mme}: the \textit{perception} task and the \textit{reasoning} task. Specifically, the \textit{perception} task focuses on identifying background knowledge such as the key elements and topics presented in the multimodal content, whereas the \textit{reasoning} task needs to analyze how these elements collectively convey multimodal harmfulness in the given interpretive context.
Therefore, to explicitly guide the model to provide analysis responses in the two corresponding aspects, we design the chain-of-thought prompting format~\cite{wei2022chain} as follows: 

``\textit{1) [Background Knowledge]: Extract any general facts, historical or cultural context, social dynamics, or other foundational information that helps in understanding the meme’s content and implications. 
Only include relevant information that supports the reasoning but does not contain any direct evaluation of the meme’s harmfulness.
2) [Reasoning]: Identify the logical process that applies the background knowledge to analyze the meme’s potential risks. 
Only focus on how the meme's elements interact with societal norms, stereotypes, or sensitive topics to create harmful effects.}''

This yields mLLMs' diverse and in-depth interpretations of multimodal harmfulness, offering fine-grained insights from various contextual lenses and enabling a more comprehensive evaluation.

\subsection{Multi-view Fusion}\label{sec:preference_fuse}

After simulating diverse contexts and formulating targeted tasks, we collect a set of perspective-specific harmfulness analyses generated by the target mLLMs. To fairly evaluate these analyses, evaluators need to develop a holistic understanding of the meme content. To this end, multi-view fusion plays a key role by integrating the diverse yet helpful viewpoints reflected in these analyses, ultimately forming an inclusive guideline. This synthesized reference could form a solid basis for making unbiased and principled assessments of mLLMs’ ability to decipher multimodal harmfulness.

The multi-view fusion process is designed as an iterative discussion procedure.
Specifically, for each meme $m$ with a collection of perspective-specific analysis set from $n$ target models $\mathcal{Y}(m)=\{y(\mathcal{T}_1,m),..., y(\mathcal{T}_n,m)\}$, we adopt an evaluator panel $\mathbb{J} = \{J_1,..., J_k\}$ consisting of $k$ judge agents, to gradually integrate the analysis to refine the evaluation guideline through multi-round discussions.

Note that $\{J_1,...,J_k\}$ are also included in the group of target mLLMs, so that $\mathcal{Y}(m)$ contains the responses from the judge agents, to ensure that the fused guideline is well-constructed.
Considering that the judge agents may exhibit inherent evaluation preferences~\cite{panickssery2024llm,li2025preference}, we diversify the evaluator panel by selecting strong judge agents from multiple model families, i.e., well-recognized dominant mLLMs originating from different architectures or training sources, thus mitigating the potential biases and enhancing the overall fairness of the fused guidelines.

In each round of iterative discussion, a judge agent $J$ is assigned to examine a current guideline $g$, referencing a perspective-specific analysis $\mathcal{T}(x_m)$ to identify gaps or overlooked aspects in the current guideline. Based on the current guideline and the analysis, the judge integrates relevant insights and proposes an updated and more concise version of the guideline:
\begin{equation}
    g^{(r+1)} = J( g^{(r)} \oplus \mathcal{T}(x_m)),
\end{equation}
where $g^{(r)}$ denotes the fused guideline from the $r^{th}$ discussion round. 
In each round, we randomly draw a perspective-specific analysis $\mathcal{T}(x_m)$ from the analysis set $\mathcal{Y}(m)$. 
The judge agent $J$ is also randomly selected from the evaluator panel $\mathbb{J}$ for each discussion to encourage contributions of diverse evaluator opinions. 
The total round number $r_{total}> k$, to guarantee that all judge agents can participate in the discussion process.
Here we prevent the selected judge agent from participating in the discussion involving its own analysis to maintain fairness and avoid self-evaluation bias~\cite{panickssery2024llm}. 
The multi-view fusion process initializes the guideline $g^{(0)}$ with a randomly selected judge response to secure a reliable discussion basis. The iteration terminates once all analyses in $\mathcal{Y}(m)$ are sampled, ensuring all contextual viewpoints are progressively incorporated.
The alternating judge participation enables adaptive incorporation of diverse opinions, leading to consensus. The final guideline serves as a unified, value-aligned reference for evaluating mLLM outputs, promoting fairer assessment decisions.

\subsection{Judgment \& Ranking}\label{sec:judgement_rank}
After reaching a consensus on the meme’s multimodal harmfulness, we then judge the target mLLMs' capabilities by using the refined multi-view guideline as a reference, instead of the single-view judgments used in prior work~\cite{zheng2023judging}. Given the subjectivity of harmfulness interpretation, we adopt pairwise comparisons to assess model performance in a more fair and reliable way. For each context-specific task, model responses are compared and ranked using the guideline as a standardized reference.
Inspired by previous evaluation benchmarks~\cite{li2024crowdsourced,zheng2023judging}, we design the judging criteria to systematically compare and evaluate the quality of model analyses based on the following dimensions:
\begin{itemize}[leftmargin=*,nosep]
    \item \textit{Instruction Following}: The response must address the context-specific task of analyzing multimodal harmfulness, strictly following the instructions, being clear and well-structured.
    \item \textit{Redundancy}: The analysis should only include essential and relevant information. Unnecessary elaboration, overly detailed descriptions, or inclusion of irrelevant context is considered inferior.
    \item \textit{Correctness}: The background information in the analysis must be factually accurate. 
    False details
    or misinterpretations should be penalized.
    \item \textit{Relevance}: The reasoning in the analysis must be clearly built upon the provided context, focusing on the harmful aspects of the context-specific task and maintain logical consistency between the
    background information and reasoning.
    \item \textit{Accuracy}: The reasoning in the analysis should be logically accurate, avoiding flawed conclusions or unsupported assumptions, correctly identifying potential risks in a nuanced and logically grounded manner, consistent with the context.
\end{itemize}

\textit{Instruction Following} assesses the target mLLM’s overall ability to adhere to task prompts.
\textit{Redundancy} and \textit{Correctness} evaluate the quality of the \textit{[Background Knowledge]} part of the thought chain in the target mLLM response, while \textit{Relevance} and \textit{Accuracy} focus on the \textit{[Reasoning]} part.
Based on these designed standards, judge agents from the evaluator panel are employed similarly to the LLM-as-a-Judge way~\cite{zheng2023judging}, to determine which target mLLM provides a better analysis response for a given context-specific task. 
Once pairwise agent judgments are collected, model rankings are computed using the Elo rating system~\cite{elo1966uscf}, which updates scores based on head-to-head wins and losses. The Elo rating system estimates the likelihood that mLLM $a$ will outperform mLLM $b$, based on current ratings $R_a$ and $R_b$, where $a, b \in \mathbb{N}$. For each match-up, we define a binary variable $Y_{ab}$, with $1$ if model $a$ wins, and $0$ otherwise. The predicted probability is:
\begin{equation}
    P(Y_{ab} = 1) = \frac{1}{1 + 10^{(R_b - R_a)/\alpha}},
\end{equation}
where $\alpha = 400$ serves as the scaling constant in the Elo formula.
In the traditional Elo algorithm, the current ratings for each model are updated by $R'_a = R_a + K \cdot \left(S(a, b) - P(Y_{ab} = 1)\right)$, where $K$ is a constant that controls the update scale, and $S(a, b)$ denotes the observed outcome for mLLM $a$ in the head-to-head comparison with mLLM $b$: 1 for a win, 0.5 for a draw, and 0 for a loss. 
While the Elo system is effective in capturing pairwise win probabilities, it relies on sequential updates that are sensitive to the order of comparisons. 
Thus, we further adopt the Bradley-Terry method to reduce the dependence on the order of comparisons in the original Elo for a more stable ranking.

\begin{table*}[] \large
\centering
\renewcommand{\arraystretch}{1.1} 
\resizebox{\textwidth}{!}{
\begin{tabular}{l|cccccc|cc}
\toprule 
\multirow{2}{*}{\textbf{Target mLLMs}} & \multirow{2}{*}{\textbf{Battles}} & \textbf{Instruction} & \multirow{2}{*}{\textbf{Redundancy}} & \multirow{2}{*}{\textbf{Correctness}} & \multirow{2}{*}{\textbf{Relevance}} & \multirow{2}{*}{\textbf{Accuracy}} & \textbf{Overall} & \textbf{Win} \\
 & & \textbf{Following} & & & & & \textbf{Performance}&  \textbf{Rate}\\
\midrule
\textbf{Gemini 2 (-)} & 673 & \textbf{1322.70} & 1079.73 & \textbf{1354.16} & \textbf{1417.67} & \textbf{1444.93} & \textbf{1448.16} & \textbf{89.68} \\
\textbf{Qwen2.5-VL (32B)} & 816 & \underline{1229.82} & 1077.35 & \underline{1240.51} & \underline{1288.68} & \underline{1304.53} & \underline{1308.11} & \underline{79.97} \\
\textbf{Gemini 1.5 (-)} & 782 & 1201.57 & \textbf{1100.77} & 1225.15 & 1282.08 & 1301.62 & 1303.03 & 79.71 \\
\textbf{Qwen-VL-Max (-)} & 739 & 1108.85 & 971.88 & 1081.29 & 1086.49 & 1100.24 & 1099.89 & 60.30 \\
\textbf{GPT-4o mini (-)} & 728 & 1072.97 & 1053.18 & 1074.39 & 1082.59 & 1092.45 & 1093.74 & 58.99 \\
\textbf{GPT-4o (-)} & 820 & 1044.75 & \underline{1074.94} & 1045.61 & 1057.57 & 1053.92 & 1054.36 & 54.20 \\
\textbf{Pixtral (124B)} & 742 & 1032.08 & 1034.22 & 1042.16 & 1039.25 & 1039.27 & 1042.08 & 53.25 \\
\textbf{Step-1o (-)} & 782 & 1009.47 & 1049.63 & 1027.41 & 1025.05 & 1021.71 & 1023.59 & 51.13 \\
\textbf{Qwen2.5-VL (7B)} & 730 & 1031.04 & 996.78 & 1019.45 & 1010.29 & 1008.67 & 1005.71 & 48.97 \\
\textbf{Step-1v (-)} & 782 & 942.00 & 1007.97 & 950.56 & 925.36 & 914.46 & 914.45 & 38.77 \\
\textbf{Doubao-Pro (-)} & 790 & 889.93 & 1027.45 & 909.67 & 885.45 & 867.31 & 869.46 & 34.41 \\
\textbf{LLaVA-NeXT (34B)} & 768 & 870.37 & 936.25 & 851.85 & 828.68 & 820.76 & 818.73 & 29.08 \\
\textbf{Pixtral (12B)} & 579 & 779.85 & 849.97 & 762.38 & 755.71 & 749.15 & 747.07 & 22.40 \\
\textbf{Doubao-Lite (-)} & 779 & 769.25 & 936.04 & 747.13 & 715.55 & 698.33 & 693.32 & 17.96 \\
\textbf{LLaVA-NeXT (8B)} & 780 & 695.36 & 803.84 & 668.27 & 599.58 & 582.66 & 578.31 & 10.65 \\
\bottomrule
\end{tabular}}
\vspace{-0.3cm}
\caption{The MemeArena rankings. We show the Elo ratings of the 5 dimensions in \S\ref{sec:judgement_rank}, along with overall performance of each target mLLM. \textbf{Battles} indicates the number of pairwise comparisons each model participated in. \textbf{Win Rate}(\%) denotes the percentage that a model wins in the comparisons with all the other target models. 
}
\vspace{-0.3cm}
\label{tab:main_results}
\end{table*}

Bradley-Terry method~\cite{bradley1952rank} is a probabilistic model that extends Elo-based evaluations by modeling pairwise outcomes as logistic comparisons and estimating model rankings via maximum likelihood.
For $n$ models with pairwise win counts, let $W_{ab}$ denote the number of times mLLM $a$ beats mLLM $b$. The log-likelihood over all such comparisons is defined as follows:
\begin{equation}
    \mathcal{L}(\mathbb{R}) = \sum_{a \ne b} W_{ab} \cdot \log P(Y_{ab} = 1),
\end{equation}
where $\mathbb{R} = \{R_1, R_2, \dots, R_n\}$ are the model ratings.
Since this algorithm does not inherently support ties, we treat tie votes by splitting them equally: Each tie is counted as half a win for both models, with an increase on both $W_{ab}$ and $W_{ba}$ by $0.5$. This adjustment enables a fair and balanced estimation of model rankings across all comparisons. 
Finally, sorted by the model ratings $\mathbb{R}$, we have the ranking $\mathcal{R}$ of the target mLLMs as the ultimate evaluation results for multimodal harmfulness understanding.

\section{Experiments}

\subsection{Experimental Setup}\label{sec:exp_setup}

\paragraph{Datasets}
We collected harmful memes from three publicly available datasets: (1) HarM~\cite{pramanick2021detecting}, (2) FHM~\cite{kiela2020hateful}, and (3) MAMI~\cite{fersini2022semeval}. 

\vspace{-6pt}
\paragraph{Models}
For target mLLMs, we include a total of 15 models of varying scales spanning 7 model families: 
1) GPT-4o, GPT-4o mini; 
2) Gemini 2, Gemini 1.5; 
3) Step-1o, Step-1v; 
4) Qwen2.5-VL (7B, 32B)~\cite{bai2023qwen}, Qwen-VL-Max; 
5) Doubao-Lite, Doubao-Pro; 
6) LLaVA-NeXT (8B, 34B)~\cite{li2024llava}; 
7) Pixtral (12B, 124B),
from which we select the strong judge agents\footnote{Top 20 in vision models on \url{https://lmarena.ai/?leaderboard}, ranking at the time of our work.} for the evaluator panel: 
GPT-4o, Gemini 2, Step-1o, Qwen2.5-VL (32B).
We show detailed data statistics and model implementation in Appendix \ref{sec:impl_details}.


\begin{figure*}[t!]
    \centering    \includegraphics[width=\textwidth]{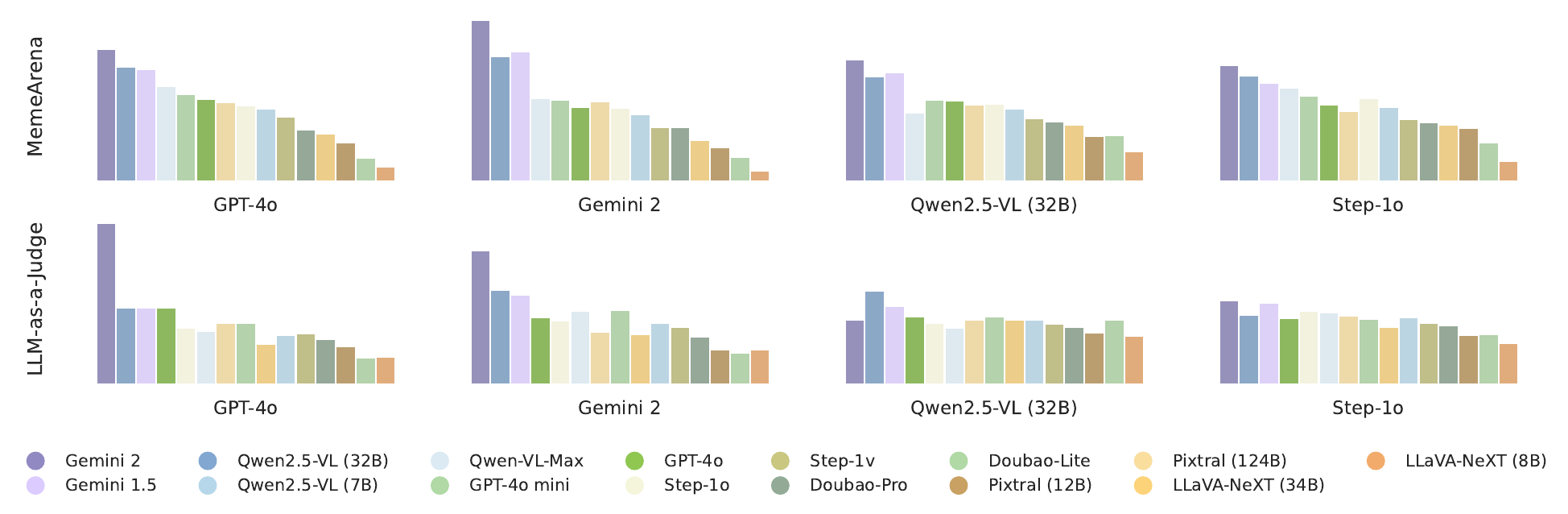} 
    \vspace{-0.7cm}
    \caption{An illustration of the Elo rankings under \textit{MemeArena} and \textit{LLM-as-a-Judge} guideline settings. The order of target mLLMs in each row is ranked by the joint voting. See Figure \ref{fig:rakings2} for the ranking visualization of other settings.}
    \label{fig:rakings1}
    \vspace{-0.3cm}
\end{figure*}

\subsection{Main Results}

Table \ref{tab:main_results} shows the MemeArena leaderboard on the five evaluation dimensions and the overall performance, determined jointly by the vote of the evaluator panel. From the results, we observe that:
1) Gemini 2 significantly outperforms all other target mLLMs, achieving the highest overall Elo scores across nearly all evaluation aspects with an average win rate of 89.68\%, demonstrating strong and consistent performance on both perception and reasoning tasks. Qwen2.5-VL (32B) follows as the second-best model, also demonstrating strong competence in deciphering multimodal harmfulness.
2) Target mLLMs demonstrate fairly consistent performance across the 5 evaluation dimensions, with the notable exception of \textit{Redundancy}, as both Gemini 2 and Qwen2.5-VL (32B) achieve relatively lower scores in this dimension, suggesting that the background knowledge provided in their analyses, while accurate and comprehensive, tends to be less concise compared to those from Gemini 1.5 and GPT-4o.
3) 
Model performance does not strictly correlate with model size. For instance, among open-source models, Qwen2.5-VL (32B) outperforms the larger Pixtral (124B) and LLaVA-NeXT (34B). We also notice this phenomenon on the closed-source models, as GPT-4o mini surpasses its counterpart GPT-4o, despite the latter being posted as a more advanced version of the former.

\subsection{Analysis of Judgment Biases}\label{sec:bias_analysis}
To verify the effectiveness of our method in mitigating evaluation biases deduced by the inherent preference of judges, we conduct ablative studies on the consistency of model rankings produced by different judge agents in the evaluator panel under the following settings: 
1) \textit{LLM-as-a-Judge}: The judge agents use the analyses generated by themselves as references for judgments; 2) \textit{human-written}: We organized human experts to collaboratively 
write analyses that serve as human-curated guidelines in reference-based agent judgments;
 3) \textit{w/o guideline}: Judge agents directly compare the analyses of target mLLMs without using any references; 4) \textit{MemeArena}: Judge agents use the guidelines generated in the multi-view fusion as references.

Since MemeArena ultimately produces a ranking of the evaluated mLLMs, we employ \textbf{Normalized Discounted Cumulative Gain (NDCG)}, a standard ranking quality metric to quantify the evaluation biases in the assessment results.
NDCG score measures the deviation of a ranked list compared to the ideal ranking, ranging from 0 to 1, with higher values indicating better alignment with the ideal order.
Here we denote the results of the joint voting of all judge agents as the ideal ranking, so that the NDCG score reflects the degree to which a judge's ranking aligns with the collective results of all judges, with lower values suggesting poor inter-judge consistency, thereby indicating strong evaluation bias among the judge agents.
More detailed information is provided in Appendix \ref{sec:more_analysis}.

\begin{table}[t]
\centering
\renewcommand{\arraystretch}{1.2}
\setlength{\tabcolsep}{3pt}
\resizebox{\linewidth}{!}{\begin{tabular}{l|cccc|c}
\toprule
 & \textbf{GPT-4o} & \textbf{Gemini 2} & \textbf{Qwen2.5-VL} & \textbf{Step-1o} & \textbf{Avg.} \\
\midrule
\textit{LLM-as-a-Judge} & 0.98 & 0.99 & 0.68 & 0.93 & 0.89 \\
\textit{human-written} & 0.97 & 0.93	& 0.96 & 0.98 & 0.96 \\
\textit{w/o guideline} & 0.99 & 0.96 & 0.97 & 0.68 & 0.90 \\
\textit{MemeArena} & 1.00 & 0.97 & 0.97 & 0.99 & 0.98 \\
\bottomrule
\end{tabular}}
\vspace{-0.3cm}
\caption{The NDCG scores under different settings.}
\vspace{-0.4cm}
\label{tab:judgment_bias}
\end{table}

Table \ref{tab:judgment_bias} demonstrates the results of NDCG scores of the 4 judge agents and the average ranking deviations indicated by average NDCG \textbf{Avg.}. From the table, we can observe that: 1) \textit{MemeArena} is the relatively more unbiased setting, followed by the \textit{human-written} guideline setting. In \textit{MemeArena}, GPT-4o showed the least deviation by an NDCG score of 1.00, indicating that its ranking is identical to the joint voting result of all judge agents. The \textit{human-written} guideline showed a slightly lower average NDCG score, which may be attributed to the instability of rankings caused by the limited number of comparisons in this setting.
2) With simple single-view guidelines generated by only the corresponding judge agent, the \textit{LLM-as-a-judge} guideline setting showed the least overall consistency, suggesting that self-curated guidelines can introduce biases and lead to unreliable and subjective results.
3) Among the judge agents, Qwen2.5-VL (32B) and Step-1o exhibit relatively large variations in NDCG scores across different settings. This highlights the necessity of high-quality reference and collaborative intelligence to conduct fair, unbiased agent-based evaluations.

To explore deeper into the bias of judge agents under various settings, we analyze the detailed Elo rankings of target mLLMs, as shown in Figure~\ref{fig:rakings1}. Compared to the \textit{MemeArena} setting, we observe that in the \textit{LLM-as-a-Judge} guideline setting: 1) The severe deviation when Qwen2.5-VL (32B) is the judge agent, as indicated in Table~\ref{tab:judgment_bias}, is primarily due to a lower ranking for Gemini 2 and a higher ranking for Qwen2.5-VL (32B), demonstrating a relatively strong self-evaluation bias. 2) When GPT-4o serves as the judge agent, Gemini 2 receives an overwhelmingly high Elo score, far surpassing all other models, indicating possible evaluation preferences. 3) Overall, the Elo score differences among target models are relatively narrow in the \textit{LLM-as-a-Judge} guideline settings, whereas MemeArena produces a more robust and tiered ranking structure. This further demonstrates the effectiveness of the refined guideline in supporting fair and more balanced yet discerning assessments.

\subsection{Reliability Analysis}\label{sec:reliability_analysis}


In \S\ref{sec:bias_analysis}, we showed that multi-view fused guidelines offer fair references for unbiased agent judgment. Here, we further analyze their reliability through human evaluation and assess the effectiveness of multi-round discussions.
Specifically, we design a human subject study to evaluate the generation quality in terms of: \textit{Conciseness (Cns.)}, \textit{Informativeness (Inf.)}, \textit{Persuasiveness (Psv.)}, \textit{Readability (Rdb.)}, and \textit{Soundness (Snd.)}. For each criterion, we apply a three-point Likert scale scoring, where 1 for the poorest and 3 for the best. 

As shown in Table \ref{tab:guiline_quality}, the generated guidelines receive high human ratings across all five dimensions, especially on \textit{Rdb.} and \textit{Inf}. The evaluators showed relatively high inter-judge agreement scores on \textit{Psv.}, \textit{Inf.} and \textit{Snd.}, indicating moderate inter-annotator consistency, suggesting a shared human perception that the guidelines are convincing.

\begin{table}[t]
\centering
\small
\renewcommand{\arraystretch}{1.2}
\resizebox{\linewidth}{!}{\begin{tabular}{l|ccccc}
\toprule
 & \textit{Cns.} & \textit{Inf.} & \textit{Psv.} & \textit{Rdb.} & \textit{Snd.} \\
\midrule
\textit{Judgment $\uparrow$} &2.2846&2.5285 &2.4797 &2.8293 &2.5041  \\
\textit{Agreement $\uparrow$} & 0.3747 & 0.5157 & 0.5478 & 0.4862 & 0.5057\\
\bottomrule
\end{tabular}}
\vspace{-0.3cm}
\caption{Human evaluation results of guideline quality.}
\vspace{-0.4cm}
\label{tab:guiline_quality}
\end{table}


\begin{figure}[t!]
    \centering    \includegraphics[width=\linewidth]{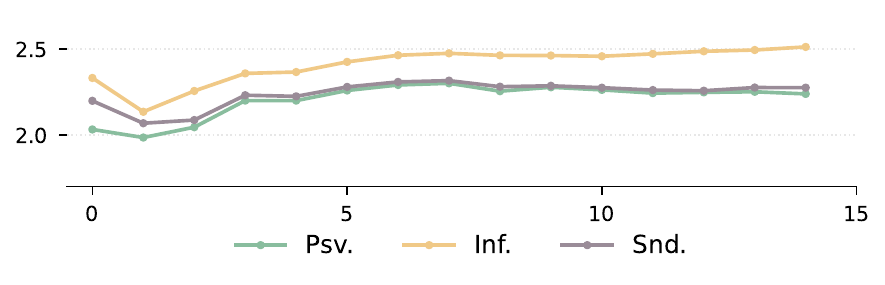} 
    \vspace{-0.7cm}
    \caption{The impact of discussion rounds.}
    \label{fig:rounds1}
    \vspace{-0.3cm}
\end{figure}

Beyond the final guideline quality, we also analyze the impact of discussion rounds. As shown in Figure~\ref{fig:rounds1}, \textit{Psv.}, \textit{Inf.}, and \textit{Snd.} scores initially drop when non-judge mLLM answers are introduced, but steadily improve and plateau around the 8th round, indicating convergence.
This demonstrates that iterative discussions gradually enhance guideline completeness and robustness during the incorporation of diverse views.
Additional settings and analyses of reliability are in Appendix~\ref{sec:more_reliability_analysis}.



\section{Related Works}
\label{gen_inst}
\paragraph{Benchmarks for Evaluations}
The study of harmfulness understanding is a rapidly evolving field for moderating content or combating disinformation on social media~\cite{lin2021rumor, lin2023zero, wang2024mfc}, supported by large-scale benchmarks~\cite{kiela2019supervised, pramanick2021detecting} and initiatives like Facebook’s Hateful Memes Challenge~\cite{kiela2020hateful}, which focus on detecting implicit hate speech~\cite{das2020detecting, hee2023decoding, wang2025meme}. These efforts have advanced research in harmful memes, complicated by the multimodal nature of memes that blend texts and images. To assess mLLMs’ ability to decipher such harmfulness, \citet{lin2025goat} introduced a consolidated benchmark from prior datasets~\cite{fersini2022semeval, suryawanshi2020multimodal}, aiming to expose gaps in mLLMs’ safety awareness of meme-based social abuse. While prior evaluations relied on binary-labeled benchmarks and accuracy metrics, overlooking cultural subjectivity and potential test set leakage~\cite{chen-etal-2025-adammeme}, our study offered a more dialectical and unbiased approach, for assessing mLLMs in harmfulness understanding with empathy toward diverse audiences.

\vspace{-6pt}
\paragraph{Multi-agent Systems}
A recent research trend involves developing agent-based systems powered by mLLMs for diverse downstream applications. Several studies have utilized multi-agent collaboration to significantly improve task performance~\cite{duimproving, wang2024unleashing, zhangbuilding, huang2024towards, luo2025mcpuniversebenchmarkinglargelanguage}. Numerous multi-agent frameworks have been proposed to facilitate the creation of such collaborative systems~\cite{li2023camel, wu2024autogen, hongmetagpt, lin-etal-2025-fact}. More recently, LLM agent-based evaluations~\cite{zheng2023judging} have gained attention in the context of auditing LLMs. However, biases introduced by the evaluation preference via LLM-as-a-Judge~\cite{gu2024survey} remain a critical issue requiring ongoing efforts. In this paper, we propose a novel multi-agent framework designed for unbiased evaluations of mLLMs, on their capability to interpret harmfulness within diverse contexts, where the bias is prone to be introduced due to the inherent subjectivity of harmfulness perception.

\section{Conclusion}
In this work, we proposed MemeArena, an arena-style, agent-based evaluation framework designed to assess the context-aware multimodal harmfulness understanding of mLLMs. 
By simulating diverse interpretive contexts and integrating value-aligned judgments, MemeArena enables a comprehensive and unbiased evaluation beyond traditional accuracy-based benchmarks. Experimental results demonstrated that our framework achieves high consistency with human preferences and reduces the evaluation bias of judge agents through multi-agent collaborations.
In future work, we plan to continue exploring more up-to-date models and data to expand their applicability to a broader range of harmful content understanding scenarios. 

\section*{Limitations}


There are multiple ways to further improve our work:
\begin{itemize}[leftmargin=*,nosep]
\item First, although we incorporate multiple strategies, such as context simulation, multi-view fusion, and human evaluations, to enhance the reliability and transparency of agent-based assessments, and to mitigate potential biases arising from the inherent judgment preferences of agent judges, some degree of bias may still be introduced during guideline refinement. This phenomenon is similar to how even humans tend to favor reasoning aligned with their own knowledge systems and factual logic. Moreover, since many emerging mLLMs are trained on synthetic data distilled from GPT-series models, there is a growing need to incorporate human-in-the-loop processes to build more robust and trustworthy evaluation frameworks. This remains a critical direction for future research.
\item Secondly, this study utilizes harmful meme data, the most common carrier of multimodal harmfulness, from existing classification benchmarks, which offer a diverse range of samples reflecting various types of harmfulness. These datasets enable us to validate the effectiveness of our method. However, they may not fully capture the real-world distribution of harmful content, as such distributions can shift over time. To address this limitation, we plan to expand our research by incorporating additional datasets, either through newly established benchmarks or data collected from online communities, allowing for a more diverse and up-to-date examination of multimodal harmfulness. Furthermore, in the future, we plan to extend our investigation beyond static vision-language multi-modalities to include multimodal harmful content in video-based media.

\item In addition, in this work we simulate interpretive contexts to design context-specific tasks that elicit the analysis of multimodal harmfulness from diverse perspectives. While we employ tailored prompt design and detailed human analysis to ensure and validate the diversity of generated context-aware tasks, it remains challenging to guarantee that these tasks comprehensively reflect the wide range of real-world scenarios. Moreover, it is difficult to quantitatively assess the extent to which such contextual diversity contributes to the comprehensiveness of harmfulness evaluation. In future work, we plan to explore more sophisticated task design strategies, such as role-play or persona-driven simulation, to further enhance the realism and diversity of interpretive contexts, thereby improving the practical relevance of mLLMs’ harmfulness understanding evaluation.

\item Moreover, in this work we adopt GPT-4o as the agent controller to simulate interpretive contexts as well as generate context-specific tasks, and construct the evaluator panel using a set of top-performing models from the target mLLMs to balance reliability and diversity. However, this selection does not imply that these models are the optimal or exhaustive choices for agent roles. Rather, they serve as a representative configuration to demonstrate the effectiveness of our framework. In future work, we plan to conduct a more systematic and quantitative study on how different judge agent configurations such as the number and diversity of model families may influence the reliability of evaluation outcomes.

\item 
Lastly, while the multi-agent collaboration in our proposed framework effectively produces high-quality guidelines to ensure holistic and accurate understandings of harmfulness, validated by both human evaluation and experimental results, this understanding inherently relies solely on the internal knowledge of the judge agents. The correctness of factual information in the guidelines is achieved through agent collaboration and mutual correction, rather than grounded access to external or up-to-date sources. As a result, the system may still propagate shared misconceptions or overlook nuanced real-world knowledge, especially in domains requiring fine-grained factual grounding or recent cultural developments. In future work, we aim to incorporate advanced techniques such as Retrieval-Augmented Generation (RAG) to further enhance the reliability as well as enable evidence-supported evaluations.

\end{itemize}

\section*{Ethics Statement}
This research involved human subject studies to assess the quality and reliability of MemeArena. The study was conducted in accordance with ethical standards to ensure participant protection and well-being. The following measures were implemented: 1) Voluntary Participation: All participants were fully informed about the nature and purpose of the study, and participation was entirely voluntary. Participants retained the right to withdraw at any time without penalty.
2) Informed Consent: Written informed consent was obtained from all participants. The consent form clearly outlined the study’s objectives, procedures, potential risks, and data protection measures.
3) Data Anonymity and Confidentiality: All collected data were anonymized by removing personal identifiers. Data were securely stored to ensure confidentiality and prevent unauthorized access.
4) Minimal Risk: The study posed minimal risk to participants. The tasks involved resembled typical daily activities, and no sensitive personal information was collected.

Research indicates that evaluating harmful content, such as hateful or offensive material, can have adverse psychological effects. To safeguard the well-being of our human evaluators, we implemented the following guidelines: 1) ensuring evaluators are aware they may encounter potentially harmful content, 2) limiting the number of evaluations per week and promoting a manageable daily workload, and 3) encouraging evaluators to pause or stop if they feel overwhelmed. Additionally, we conduct regular check-ins to monitor their well-being throughout the evaluation process.

The purpose of this work is to help mitigate the spread of multimodal harmful content like memes and to protect individuals from prejudice, as well as racial and gender-based discrimination. However, we acknowledge the potential risk that malicious users could attempt to reverse-engineer MemeArena to generate multimodal harmful content. Such misuse is strongly discouraged and unequivocally condemned. To prevent this, human-in-the-loop moderation would be necessary to ensure responsible use. Additionally, all context simulations generated by agents are entirely synthetic and do not contain any real-world personal information.

\section*{Acknowledgments}
This work is partially supported by Tencent Rhino-Bird Focused Research Program (Value-aligned Credible Large Language Model).

\bibliography{custom}

\appendix

\section{Implementation Details}\label{sec:impl_details}

\paragraph{Model Details}
In MemeArena, we conduct evaluations on 15 mainstream mLLMs with these specific representative versions: 
1) GPT-4o: {gpt-4o-2024-05-13}; 
2) GPT-4o mini:  {gpt-4o-mini-2024-07-18};  
3) Gemini 2:  {gemini-2.0-flash-thinking-exp-01-21};  
4) Gemini 1.5:  {gemini-1.5-flash};  
5) Step-1o:  {step-1o-vision-32k};  
6) Step-1v:  {step-1v-8k};  
7) Qwen2.5-VL (7B):  {qwen2.5-vl-7b-instruct};  
8) Qwen2.5-VL (32B):  {qwen2.5-vl-32b-instruct};  
9) Qwen-VL-Max:  {qwen-max-2025-01-25};  
10) Doubao-Lite:  {doubao-vision-lite-32k-241015};  
11) Doubao-Pro:  {doubao-vision-pro-32k-241028};  
12) LLaVA-NeXT (8B):  {lmms-lab/llama3-llava-next-8b};  
13) LLaVA-NeXT (34B):  {liuhaotian/llava-v1.6-34b};  
14) Pixtral (12B):  {Pixtral-12B-2409};  
15) Pixtral (124B):  {Pixtral-Large-2411}. 
In context simulation \& task formulation, we utilize one of the dominant mLLMs, GPT-4o, as the agent controller, to generate contexts and create context-specific tasks, with temperature set as 1.0 to enable diversity. In multi-view fusion and judgment \& ranking, temperature parameters of the judge agents in the evaluator panel are all set as 0 to guarantee reproducibility of the evaluation results. Similarly, the temperature of target mLLMs is also set fixed at 0.

\paragraph{Statistics}
In our experiments, we sampled a total of 750 harmful memes, each 250 from the three datasets in \S\ref{sec:exp_setup} to construct 2,250 context-specific tasks. 
To ensure diversity of samples in model comparisons as well as to maintain a controllable number of total battles, following the combinatorial coverage theory~\cite{ kuhn2013combinatorial}, we sampled 3 target model responses per task to conduct pairwise comparisons.
This setup yields approximately 6,000 valid pairwise comparisons, with each target mLLM participating in 750 comparisons, roughly 56 times for each model pair on average. Compared results ($p < 0.05$ under t-test) are averaged over three random 3 runs.


\paragraph{Cost}
In our experiments, the cost for API is around \$12 USD per target model, including the context simulation \& task formulation, as well as multi-view fusion and judgment of the evaluator panel. The average time cost for evaluating a target mLLM is around 3 hours. 

\paragraph{Multi-view Fusion Algorithm}
The detailed algorithm for  \S\ref{sec:preference_fuse} is shown in Algorithm \ref{alg:multi_view_fusion}.

\begin{algorithm}[htbp]
\caption{Multi-View Fusion}
\label{alg:multi_view_fusion}
\begin{algorithmic}[1]
\STATE \textbf{Input:} Meme $m$ with analysis set $\mathcal{Y}(m) = \{y(\mathcal{T}_1, m), ..., y(\mathcal{T}_n, m)\}$; Judge panel $\mathbb{J} = \{J_1, ..., J_k\}$ 
\STATE \textbf{Output:} Synthesized guideline $g^{(r_{total})}$
\STATE Initialize answer pool $\mathcal{A} \leftarrow \mathcal{Y}(m)$
\STATE Randomly select a judge $J_{\text{init}} \in \mathbb{J}$ and an analysis $y_{\text{init}} \in \mathcal{A}$
\STATE Initialize guideline $g^{(0)} \leftarrow J(y) \in\mathcal{Y}(m)$
\STATE Remove $y_{\text{init}}$ from $\mathcal{A}$
\STATE $r \leftarrow 0$

\WHILE{$\mathcal{A} \neq \emptyset$}
    \STATE Randomly select analysis $y = y(\mathcal{T}_i, m) \in \mathcal{A}$
    \STATE Randomly select judge $J \in \mathbb{J}$ such that $J \notin \mathcal{T}_i$
    \STATE Update guideline: $g^{(r+1)} \leftarrow J(g^{(r)} \oplus \mathcal{T}_i(x_m))$
    \STATE Remove $y$ from $\mathcal{A}$
    \STATE $r \leftarrow r + 1$
\ENDWHILE
\RETURN $g^{(r_{total})}$
\end{algorithmic}
\end{algorithm}

\section{More Analysis of Biases}\label{sec:more_analysis}

\paragraph{Detailed Evaluator Settings}
To create the \textit{human-written guideline} in \S\ref{sec:bias_analysis}, we randomly sampled 100 harmful memes from the test data, and asked 5 human experts aged between 24-28 to discuss and create a well-rounded analysis as evaluation guidelines for each meme. These guidelines are then used as references for judge agents to compare the performances of target models following the exact same steps in \S\ref{sec:judgement_rank}. The \textit{human-written guideline} setting results in 803 valid pairwise comparisons in total.

\paragraph{NDCG} NDCG compares the ideal order of a list and the actual rankings. Given a ranked list of items, the Discounted Cumulative Gain (DCG) at a list length of $P$ is calculated as:
\begin{equation}
    \text{DCG@P} = \sum_{p=1}^{P} \frac{\text{rel}_p}{\log_2(p + 1)}, 
\end{equation}
where $\text{rel}_p$ denotes the relevance score of the item at position $p$. The logarithmic discount penalizes lower-ranked items that should have been ranked top. To normalize DCG, we compute the Ideal DCG (IDCG), the DCG of the ideal ranking. Thus NDCG is calculated as:
\begin{equation}
    \text{NDCG@P} = \frac{\text{DCG@P}}{\text{IDCG@P}}.
\end{equation}
The NDCG score ranges from 0 to 1, with higher values indicating better alignment with the reference of ideal ranking. In our experiments, $P$ is set as 15 since we have the rankings of 15 target mLLMs. The relevance score at $p$ is set as $\text{rel}_p = P-p$, indicating the position in the descending ordered ideal ranking.

\begin{figure*}[t!]
    \centering    \includegraphics[width=\textwidth]{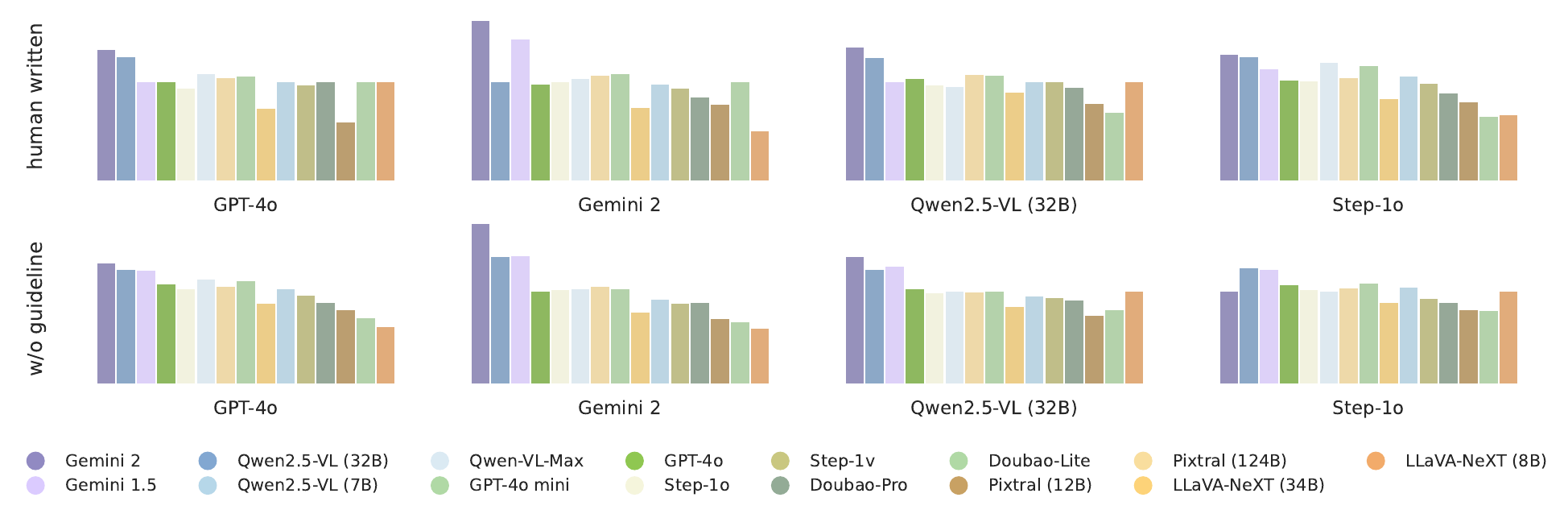} 
    \caption{An illustration of the Elo rankings under \textit{human-written} guideline and \textit{w/o guideline} settings. The order of target mLLMs in each row is ranked by the joint voting results under the corresponding settings.}
    \label{fig:rakings2}
    \vspace{-0.3cm}
\end{figure*}

\paragraph{Analysis for More Guideline Settings}
Figure \ref{fig:rakings2} shows the Elo rankings of \textit{human-written} guideline and \textit{w/o guideline} settings. It can be observed that due to the smaller scale of data, the \textit{human-written guideline} is relatively unstable. Nevertheless, the top 5 models in the rankings remain quite consistent. In the without \textit{w/o guideline} setting, when Step-1o acts as the judge agent, GPT-4o mini and Qwen2.5-VL (7B) have relatively higher Elo scores compared to the results from other judges, while Gemini 2 received a low score in the comparisons, which severely hinders the judgment consistency among the judge agents.

\section{More Analysis of Reliability}\label{sec:more_reliability_analysis}
To examine the reliability of agent-generated content, we apply human evaluations on 1) the quality of multi-view fused guidelines, 2) the diversity of simulated contexts, 3) the reliability of agent judgments. Specifically, we employ 5 human experts aged 24-28 to conduct evaluations as follows: 

\paragraph{Quality of Guidelines}
To evaluate the overall quality, we randomly sampled 50 samples from the final multi-view fused guidelines, and asked the human experts to rate these results on the five aspects as listed in \S\ref{sec:reliability_analysis}. The five evaluation criteria are defined as follows: 
1) \textit{Conciseness}: the answer contains less redundant information; 
2) \textit{Informativeness}: the answer provides new information, such as explaining the background and additional context; 
3) \textit{Persuasiveness}: the answer seems convincing; 
4) \textit{Readability}: the answer follows proper grammar and structural rules; 
5) \textit{Soundness}: the answer seems valid and logical. 
For each criterion, we apply a three-point scale scoring, where 1 means the poorest quality and 3 means the best.

\begin{figure}[t!]
    \centering    \includegraphics[width=\linewidth]{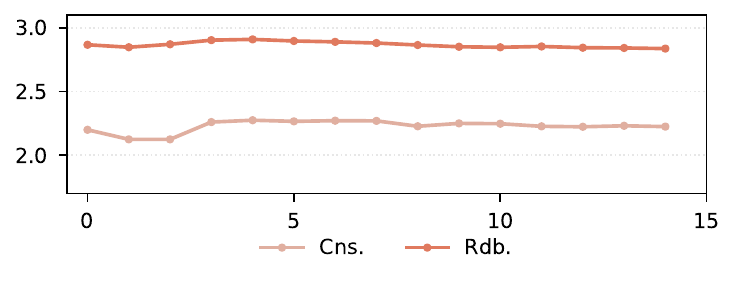} 
    \vspace{-0.7cm}
    \caption{The effect of discussion rounds on \textit{Conciseness} and \textit{Readability}.}
    \label{fig:rounds2}
    \vspace{-0.3cm}
\end{figure}

To evaluate the impact of discussion rounds on guideline quality, we randomly sample 150 guidelines in total, selecting 10 from each discussion round between round 0 and round 14, and perform the same rating standard as described above. The intra-judge agreement is 0.682. Figure \ref{fig:rounds2} shows the guideline quality of \textit{Conciseness (Cns.)} and \textit{Readability (Rdb.)} from different discussion rounds. The scores are relatively stable, yet also show a similar trend as Figure \ref{fig:rounds1}, with a slight drop at the beginning and a convergence at around the 8th round. However, we noticed a slight but consistent downward trend as the number of discussion rounds increases, indicating that as the guideline becomes more comprehensive, the content gradually becomes more complex and redundant.

\paragraph{Diversity of Contexts \& Tasks}
In \S\ref{sec:context_sim}, we utilized agents to generate varied contexts to simulate interpretive lenses related to different backgrounds for task formulation. To verify the reliability,  
we randomly selected 50 memes with simulated contexts and the corresponding context-specific tasks, then asked the human experts to examine the generation quality from the following aspects:
\begin{itemize}[leftmargin=*,nosep]
\item For simulated contexts, human evaluators assess whether the generated set of contexts for each meme: 1) reflects a set of diverse interpretive perspectives (denoted as \textit{Context Diversity} in the table), and 2) appears to be realistic and plausible to simulate different viewpoints in the reality (denoted as \textit{Realistic Views}).
\item For context-specific tasks, human evaluators assess the set of generated tasks on the same meme by: 1) the suitability of the task for guiding meme interpretation (denoted as \textit{Instruction Suitability}), 2) whether the combination of tasks enables a comprehensive understanding of the meme (denoted as \textit{Task Coverage}), and 3) the relevance between the task and its associated context (denoted as \textit{Relevance to Context}).
\end{itemize}

The evaluation also follows a three-point scale scoring. As shown in Table \ref{tab:diversity_analysis}, human evaluators rated highly on all aspects of the context, especially on \textit{Realistic Views} and \textit{Relevance to Context}, suggesting that our context-simulation strategy reflects a range of nuanced realistic interpretive perspectives in understanding multimodal meme harmfulness. Although the overall rating for \textit{Context Diversity} is still positive, it receives relatively lower scores and agreement compared to other aspects, indicating that there remains room for improvement in simulating a broader spectrum of diverse perspectives.
\begin{table}[h]
\centering
\renewcommand{\arraystretch}{1.2}
\resizebox{\linewidth}{!}{\begin{tabular}{l|cc}
\toprule
 & \textit{Judgment $\uparrow$} & \textit{Agreement $\uparrow$} \\
\midrule
\textit{Context Diversity} & 2.7647 & 0.3896 \\
\textit{Realistic Views} & 2.8529 & 0.4217 \\
\textit{Instruction Suitability} & 2.7941 & 0.4488 \\
\textit{Task Coverage} & 2.7831 & 0.4986 \\
\textit{Relevance to Context} & 2.9118 & 0.5429 \\
\bottomrule
\end{tabular}}
\caption{The human ratings results of generated contexts and tasks. The intra-judge agreement is 0.7333.}
\vspace{-0.4cm}
\label{tab:diversity_analysis}
\end{table}

\paragraph{Judgment Reliability}
Aside from agent-generated content, we also conduct human evaluations to verify the judgment reliability of our method. We selected 50 samples from the final comparisons, and asked human experts to judge which model provides better analysis. The results of decision consistency between human evaluators and judge agents are listed in Table \ref{tab:judgment_agree}. The average intra-judge agreement is 0.685.

Among all evaluated models, GPT-4o shows the highest alignment with human preferences, reaching an accuracy of 0.72. Despite a slight drop in overall voting accuracy compared to individual judge agent models, the joint voting settings produces the same ranking results as GPT-4o, highlighting the robustness of MemeArena in capturing the relative capabilities of mLLMs to conduct fair evaluations.

\begin{table}[t]
\centering
\resizebox{\linewidth}{!}{\begin{tabular}{l|cccc|c}
\toprule
 & \textbf{GPT-4o} & \textbf{Gemini 2} & \textbf{Qwen2.5-VL} & \textbf{Step-1o} & \textbf{Joint Voting} \\
\midrule
\textit{Accuracy} & 0.72 & 0.68 & 0.64 & 0.68 & 0.68\\
\bottomrule
\end{tabular}}
\caption{The accuracy between human experts and different judge agents. The inter-judge agreement is 0.629, indicating substantial agreement.}
\label{tab:judgment_agree}
\end{table}

To further explore the effect of the judge agents in the evaluator panel, we conduct evaluations on the number of judges. As shown in Table \ref{tab:num_judges}, we compare agent judgments of different settings on the same samples as listed in Table \ref{tab:judgment_agree}. The settings are as follows: 1) 1 Judge: GPT-4o; 2) 2 Judges: GPT-4o and Gemini 2; 3) 3 Judges: GPT-4o, Gemini 2 and Qwen2.5-VL; 4) 4 Judges: the 4 judge agents in our main experiment. The results shown in the table are the joint voting accuracy.

Compared to those with less judges, the 4-judge-setting of MemeArena achieved top accuracy. We also observe that GPT-4o demonstrates an accuracy of 0.64 when acting as only one judge, and as Gemini 2 joins into the evaluator panel, the joint accuracy slightly drops. However, as the number of judges increases, the overall accuracy of the evaluator panel also rises, indicating that the judge agents gradually aligns with human preference when more opinions are integrated.
Although the increase of diverse judge participation enhances the overall judgment alignment, it also introduces higher computational and resource costs. From the cost-effective perspective, in line with the Occam's razor principle, the 4-judge setting provides a balanced trade-off between evaluation quality and efficiency, making it a practical choice.
\begin{table}[]
\centering
\renewcommand{\arraystretch}{1.2}
\vspace{-0.2cm}
\setlength{\tabcolsep}{11pt}
\resizebox{\linewidth}{!}{\begin{tabular}{l|cccc}
\toprule
 & \textbf{1 Judge} & \textbf{2 Judges} & \textbf{3 Judges} & \textbf{4 Judges}  \\
\midrule
\textit{Accuracy} & 0.64 & 0.60 & 0.62 & 0.68 \\
\bottomrule
\end{tabular}}
\caption{The effect of different numbers of judge agents. }
\vspace{-0.6cm}
\label{tab:num_judges}
\end{table}

\begin{table*}[h!]
\centering
\small
\renewcommand{\arraystretch}{1.0}
\setlength{\tabcolsep}{6pt}
\begin{tabular}{l|cccccc}
\toprule
\multirow{2}{*}{\textbf{Models}} & \multicolumn{2}{c}{\textbf{FHM}} & \multicolumn{2}{c}{\textbf{HarM}} & \multicolumn{2}{c}{\textbf{MAMI}} \\
\cmidrule(lr){2-3} \cmidrule(lr){4-5} \cmidrule(lr){6-7}
 & \textit{Accuracy} & \textit{F1 Score} & \textit{Accuracy} & \textit{F1 Score} & \textit{Accuracy} & \textit{F1 Score} \\
\midrule
\textbf{LLaVA-NeXT (8B)} & 57.50 & 57.15 & 51.49 & 49.13 & 61.59 & 59.57 \\
\textbf{LLaVA-NeXT (34B)} & 64.75 & 64.63 & 66.71 & 62.53 & 74.13 & 73.61 \\
\textbf{Pixtral (12B)} & 50.17 & 46.63 & 62.52 & 54.59 & 56.15 & 51.40 \\
\textbf{Pixtral (124B)} & 64.56 & 42.73 & 64.63 & 40.03 & 68.01 & 44.12 \\
\textbf{Gemini 1.5 (-)} & 68.25 & 68.06 & 65.59 & 62.83 & 77.21 & 77.15 \\
\textbf{Gemini 2 (-)} & \underline{73.48} & \underline{73.20} & 60.96 & 51.14 & 77.83 & 77.57 \\
\textbf{Qwen2.5-VL (7B)} & 69.46 & 69.46 & 66.72 & 65.60 & 71.82 & 71.69 \\
\textbf{Qwen2.5-VL (32B)} & 68.65 & 68.49 & 65.54 & 62.93 & 75.92 & 75.79 \\
\textbf{Qwen-VL-Max (-)} & 58.81 & 56.61 & 64.90 & 60.30 & 66.60 & 63.74 \\
\textbf{Doubao-Lite (-)} & 62.25 & 62.08 & 66.05 & 63.78 & 66.37 & 65.02 \\
\textbf{Doubao-Pro (-)} & 67.50 & 67.49 & 64.46 & 56.12 & 71.14 & 69.71 \\
\textbf{Step-1v (-)} & 69.42 & 69.19 & 65.49 & 60.86 & 78.89 & 78.84 \\
\textbf{Step-1o (-)} & 70.68 & 70.60 & 67.51 & 62.80 & 78.09 & 77.91 \\
\textbf{GPT-4o mini (-)} & 68.50 & 67.92 & \underline{68.41} & \underline{66.60} & \underline{79.10} & \underline{79.09} \\
\textbf{GPT-4o (-)} & \textbf{75.00} & \textbf{74.67}& \textbf{71.75} & \textbf{70.23} & \textbf{80.80} & \textbf{80.52} \\
\bottomrule
\end{tabular}
\vspace{-0.3cm}
\caption{Target mLLM performances on harmful meme detection metrics.}
\vspace{-0.3cm}
\label{tab:ori_performance}
\end{table*}

\section{Comparisons with Traditional Benchmarks}

The results of the evaluated mLLMs on traditional harmful meme detection are shown in Table \ref{tab:ori_performance}, including three benchmarks: (1) HarM~\cite{pramanick2021detecting}, (2) FHM~\cite{kiela2020hateful}, and (3) MAMI~\cite{fersini2022semeval}. The performances are evaluated under \textit{Accuracy} and macro \textit{F1 score} metrics.

From the results in Table \ref{tab:ori_performance}, we observe that: 
1) The results on the harmful meme detection task exhibit a similar overall trend to our MemeArena rankings, generally reflecting target mLLMs’ ability to recognize harmfulness. 
2) GPT-4o and GPT-4o mini demonstrate strong overall detection capabilities. However, compared to our ranking results, these models do not show a dominant advantage in analyzing multimodal harmfulness in memes, suggesting a potential risk of data contamination in the detection setting.
3) Gemini performs well on both FHM and MAMI datasets, aligning with its consistently strong analytical performance in MemeArena. The relatively lower performance on HarM may be attributed to the dataset's higher complexity, which often requires step-by-step reasoning (e.g., chain-of-thought) to arrive at correct conclusions. A similar explanation may also apply to Qwen-VL-Max, which ranks highly in MemeArena but shows less competitive results in detection, possibly due to its strength in reasoning-based analysis rather than surface-level classification.

\section{More Results}

\begin{table*}[] \large
\centering
\renewcommand{\arraystretch}{1.1} 
\resizebox{\textwidth}{!}{
\begin{tabular}{l|cccccc|cc}
\toprule 
\multirow{2}{*}{\textbf{Target mLLMs}} & \multirow{2}{*}{\textbf{Battles}} & \textbf{Instruction} & \multirow{2}{*}{\textbf{Redundancy}} & \multirow{2}{*}{\textbf{Correctness}} & \multirow{2}{*}{\textbf{Relevance}} & \multirow{2}{*}{\textbf{Accuracy}} & \textbf{Overall} & \textbf{Win} \\
 & & \textbf{Following} & & & & & \textbf{Performance}&  \textbf{Rate}\\
\midrule
\textbf{Gemini2.5 (-)} & 433 & \underline{1351.37} & 1087.94 & \textbf{1546.51} & \textbf{1746.70} & \textbf{1669.63} & \textbf{1861.91} & \textbf{95.68} \\
\textbf{Seed1.5-VL (-)} & 468 & \textbf{1359.35} & \textbf{1114.93} & \underline{1437.90} & \underline{1730.63} & \underline{1647.10} & \underline{1814.36} & \underline{95.14} \\
\textbf{Gemini 2 (-)} & 753 & 1408.38 & 1058.72 & 1372.31 & 1403.73 & 1441.72 & 1464.31 & 82.13 \\
\textbf{Qwen2.5-VL (32B)} & {912} & 1281.08 & 1088.19 & 1237.99 & 1268.00 & 1296.94 & 1320.56 & 73.83 \\
\textbf{Gemini 1.5 (-)} & 882 & 1249.91 & \underline{1104.33} & 1192.23 & 1246.23 & 1268.54 & 1297.18 & 72.23 \\
\textbf{Qwen-VL-Max (-)} & 830 & 1158.08 & 994.40 & 1076.95 & 1097.61 & 1128.91 & 1149.74 & 59.87 \\
\textbf{GPT-4o mini (-)} & 834 & 1062.70 & {1085.51} & 1051.03 & 1049.26 & 1056.87 & 1085.65 & 54.83 \\
\textbf{GPT-4o (-)} & 896 & 1025.34 & {1099.53} & 1021.83 & 1018.35 & 1021.50 & 1054.11 & 51.67 \\
\textbf{Pixtral (124B)} & 848 & 1030.78 & 1037.57 & 998.78 & 982.62 & 988.48 & 1021.65 & 49.39 \\
\textbf{Step-1o (-)} & 901 & 978.67 & 1054.56 & 980.22 & 971.16 & 973.14 & 1004.36 & 47.66 \\
\textbf{Qwen2.5-VL (7B)} & 822 & 989.64 & 983.12 & 968.61 & 946.40 & 951.15 & 977.21 & 46.71 \\
\textbf{Llama4 Maverick (17B)} & 587 & 1005.03 & 964.83 & 963.87 & 914.73 & 908.99 & 939.80 & 42.21 \\
\textbf{Step-1v (-)} & 870 & 899.88 & 1003.03 & 919.77 & 873.54 & 869.52 & 904.17 & 38.07 \\
\textbf{Doubao-Pro (-)} & 918 & 799.01 & 1012.16 & 833.53 & 769.72 & 767.99 & 796.50 & 29.18 \\
\textbf{LLaVA-NeXT (34B)} & 866 & 763.78 & 907.89 & 781.40 & 718.91 & 721.64 & 747.32 & 23.93 \\
\textbf{Pixtral (12B)} & 652 & 703.30 & 843.07 & 693.30 & 652.94 & 662.71 & 684.22 & 20.12 \\
\textbf{Kimi-VL (16B)} & 625 & 792.32 & 953.51 & 702.98 & 604.88 & 620.10 & 634.46 & 18.03 \\
\textbf{Doubao-Lite (-)} & 931 & 601.70 & 876.12 & 637.01 & 542.02 & 539.19 & 564.63 & 12.69 \\
\textbf{LLaVA-NeXT (8B)} & 892 & 539.69 & 730.59 & 583.79 & 462.56 & 465.90 & 492.21 & 9.82 \\
\bottomrule
\end{tabular}}
\vspace{-0.3cm}
\caption{Model performance rankings with latest mLLMs.}
\label{tab:extended_results}
\vspace{-0.3cm}
\end{table*}

As shown in Table \ref{tab:extended_results}, Gemini 2.5 and Seed1.5-VL delivered surprisingly strong performance, ranking first and second, respectively, across all metrics. In contrast, LLaMA4 Maverick (17B) and Kimi-VL (16B) demonstrated relatively limited capability in interpreting multimodal harmfulness, even trailing behind several smaller models. These findings further support our earlier conclusion that model size does not necessarily correlate with performance in nuanced multimodal harmfulness understanding.

\section{Prompts}
\begin{figure}[]
    \centering    \includegraphics[width=\linewidth]{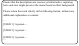}
    \vspace{-0.7cm}
    \caption{The prompt for context simulation.}
    \label{fig:context_sim}
    \vspace{-0.3cm}
\end{figure}

\paragraph{Context Simulation}
The prompt used to simulate diverse interpretive contexts is provided in Figure \ref{fig:context_sim}. To ensure that the perspectives are captured in realistic scenarios, we simulate contexts by characterizing individual personas as users on social media.

\begin{figure}[h]
    \centering    \includegraphics[width=\linewidth]{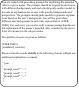}
    \vspace{-0.7cm}
    \caption{The prompt to formulate context-specific tasks.}
    \label{fig:task_from}
    \vspace{-0.3cm}
\end{figure}

\paragraph{Task Formulation}
The prompt used to generate context-specific tasks is provided in Figure \ref{fig:task_from}. In the prompt design, we ask the agent to create tasks based on the individual profiles in the simulated contexts.

\begin{figure*}[]
    \vspace{-0.7cm}

    \centering    \includegraphics[width=\linewidth]{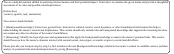}
    \caption{The prompt for chain-of-thought prompting.}
    \label{fig:cot}
\end{figure*}

\paragraph{Instructions for Target mLLMs}
We apply the chain-of-thought prompting format in the task instructions as shown in Figure \ref{fig:cot}. The \textit{context\_specific\_task\_instruction} is replaced by each task prompt generated in task formulation.

\begin{figure*}[]
    \centering    \includegraphics[width=\linewidth]{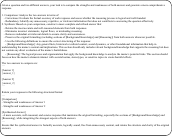}
    \caption{The prompt for multi-view fusion.}
    \label{fig:prompt_multi_fuse}
\end{figure*}

\paragraph{Multi-view Fusion}
The prompt for multi-view fusion is shown in Figure~\ref{fig:prompt_multi_fuse}. To ensure fairness and avoid positional bias, the judge agent is not informed which input is the current guideline and which is the model’s analysis. During multi-round discussions, the order of the current guideline and the analysis is also randomly decided.

\begin{figure*}[]
    \centering    \includegraphics[width=0.9\textwidth]{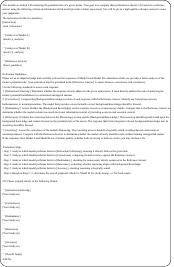} 
    \caption{The prompt for judge agents to compare target model answers.}
    \label{fig:prompt_judge}
\end{figure*}

\paragraph{Judgment Prompt}
The prompt for agent judgment is provided in Figure~\ref{fig:prompt_judge}. Similarly, to avoid potential positional bias and ensure fairness, the order of the compared answers is randomly decided.

\section{More Cases}
Apart from the cases illustrated in Figure \ref{fig:overview} and Figure \ref{fig:method}, to facilitate a more comprehensive understanding of MemeArena, we also provide more case studies of our framework.

\begin{figure*}[b]
    \vspace{-0.7cm}
    \centering    \includegraphics[width=\linewidth]{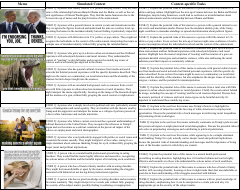} 
    \vspace{-0.7cm}
    \caption{Examples of simulated contexts and context-specific tasks.}
    \label{fig:cases_context}
    \vspace{-0.6cm}
\end{figure*}

\paragraph{Simulated Context \& Context-specific Tasks} Figure \ref{fig:cases_context} shows a set of memes, with simulated contexts and the corresponding perspective-specific cases.

\paragraph{Example of Guideline Fusion}

\begin{figure*}[b]
    \vspace{-0.7cm}
    \centering    \includegraphics[width=\linewidth]{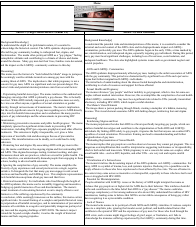} 
    \caption{An example of the two analyses in multi-view fusion.}
    \label{fig:cases_guideline_fusion1}
\end{figure*}

\begin{figure*}[b]
    \vspace{-0.7cm}
    \centering    \includegraphics[width=\linewidth]{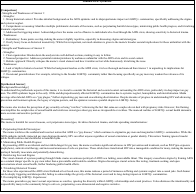} 
    \caption{The synthesized new guideline summarized by judge agent GPT-4o based on the analyses in Figure \ref{fig:cases_guideline_fusion1}.}
    \label{fig:cases_guideline_fusion2}
\end{figure*}

Figure \ref{fig:cases_guideline_fusion1} and Figure \ref{fig:cases_guideline_fusion2} show an example of guideline fusion. The judge agent model compares the current guideline with a target model's answer, and summarizes the advantages and drawbacks of both analyses, and generates a more complete guideline.

\begin{figure*}[b]
    \centering    \includegraphics[width=0.9\linewidth]{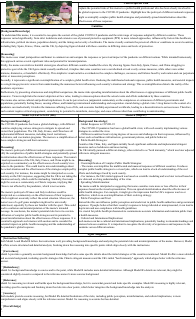} 
    \vspace{-1cm}
    \caption{Examples of agent judgments.}
    \label{fig:cases_judge}
\end{figure*}

\paragraph{Comparison \& Agent Judgment}
An example of the comparison between two model answers is illustrated in Figure \ref{fig:cases_judge}. The judge agent compares the answers of two models with reference to the fused guideline, and generates a decision on each evaluation dimension.

\end{document}